\documentclass{article}

\usepackage[final]{neurips_2024}

\usepackage[utf8]{inputenc} 
\usepackage[T1]{fontenc}    
\usepackage{hyperref}       
\usepackage{url}            
\usepackage{booktabs}       
\usepackage{amsfonts}       
\usepackage{nicefrac}       
\usepackage{microtype}      
\usepackage{xcolor}         
\usepackage{graphicx}
\usepackage{amsmath}
\usepackage{multirow}
\usepackage{amsmath} 
\usepackage{amssymb} 
\newcommand{\cmark}{\ding{51}} 
\newcommand{\xmark}{\ding{55}} 
\usepackage{graphicx}
\usepackage{hyperref}
\usepackage{adjustbox}
\usepackage{multicol}
\usepackage{multirow}
\usepackage{xcolor}
\usepackage{pifont}

\title{Deep Generative Models Unveil Patterns in Medical Images Through Vision- “Language” Conditioning}

\author{
  Xiaodan Xing\thanks{contributed equally}\\
  Bioengineering Department and Imperial-X \\
  Imperial College London\\
  London, United Kingdom \\
  \texttt{xiaodan.xing@imperial.ac.uk} \\
  \And
    Junzhi Ning\textsuperscript{*} \\
  Bioengineering Department and Imperial-X \\
  Imperial College London\\
  London, United Kingdom \\
  \And
    Yang Nan \\
  Bioengineering Department and Imperial-X \\
  Imperial College London\\
  London, United Kingdom \\
  \And
  Guang Yang \\
  Bioengineering Department and Imperial-X \\
  Imperial College London\\
  London, United Kingdom \\
  \texttt{guang.yang@imperial.ac.uk} \\
}

\begin{document}

\maketitle

\begin{abstract}
Deep generative models have significantly advanced medical imaging analysis by enhancing dataset size and quality. Beyond mere data augmentation, our research in this paper highlights an additional, significant capacity of deep generative models: their ability to reveal and demonstrate patterns in medical images. We employ a generative structure with hybrid conditions, combining clinical data and segmentation masks to guide the image synthesis process. Furthermore, we innovatively transformed the tabular clinical data into textual descriptions. This approach simplifies the handling of missing values and also enables us to leverage large pre-trained vision-language models that investigate the relations between independent clinical entries and comprehend general terms, such as gender and smoking status. Our approach differs from and presents a more challenging task than traditional medical report-guided synthesis due to the less visual correlation of our clinical information with the images. To overcome this, we introduce a text-visual embedding mechanism that strengthens the conditions, ensuring the network effectively utilizes the provided information. Our pipeline is generalizable to both GAN-based and diffusion models. Experiments on chest CT, particularly focusing on the smoking status, demonstrated a consistent intensity shift in the lungs which is in agreement with clinical observations, indicating the effectiveness of our method in capturing and visualizing the impact of specific attributes on medical image patterns. Our methods offer a new avenue for the early detection and precise visualization of complex clinical conditions with deep generative models. All codes are  \url{https://github.com/junzhin/DGM-VLC}.
\end{abstract}

\section{Introduction}
Deep generative models have traditionally served as vital tools for data augmentation in medical image analysis, enhancing the volume and quality of datasets for downstream tasks. However, the ever-increasing volume of real medical data during routine scanning and advancements in image acquisition algorithms challenge the necessity of using these models merely for data augmentation, especially when synthetic data may not match the quality of real observations. This evolution prompts a critical reassessment of the broader applications of deep generative models beyond simple data augmentation.

The application of generative models for anomaly detection through reconstruction techniques marked a significant shift \cite{schlegl2017unsupervised,han2021madgan}. By training on healthy data and inferencing on patient data, these models can highlight differences as anomalies. However, the binary nature of this patient v.s. control strategy limits its application on comparisons across multiple classes, such as categorizing patients into different age groups.

Addressing these challenges, our research proposes a novel aspect of utilizing generative models to identify pattern that correlates with clinical attributes. Our method features by a hybrid condition, including both clinical information, such as gender, age, and diagnosis results, and segmentation masks, to guide the image synthesis process. The clinical information guidance enables the generation of diverse medical image patterns, and the segmentation masks offer structural guidance to minimize bias and highlight distinctive patterns.

The pervasive issue of missing data represents a significant obstacle to our concept's implementation. Our solution involves transforming tabular clinical data into detailed textual descriptions, allowing us to bypass the challenges posed by missing values. This conversion also exploits the potential of pre-trained vision-language models to understand clinical information expressed in simple terms, such as gender and smoking status.

Another challenge in our implementation is that, unlike conventional medical report-guided synthesis, our algorithm is conditioned on clinical information with no direct visual correlation to the images. Medical reports narrate observable patterns in medical images, while clinical parameters—like age, gender, and smoking status—lack established visual representations. Thus, we explore two approaches of text fusion unit including cross-attention module and Affine transformation fusion unit to enhance the conditions, aiming to signify the conditions on generated images. 

Our experiments, conducted on a publicly available chest CT dataset, not only showcase the superior synthesis performance of our proposed framework but also highlight its effectiveness in capturing and visualizing the impacts of clinical status in medical image patterns, matching with clinical observations. 

In summary, the primary contribution of our study is a novel method that employs generative models to detect medical image patterns that are associated with clinical attributes like age, gender, and smoking history. Our technical advancements include 1) \textbf{Conversion of tabular data into text}, which addresses missing data issues and utilizes the capabilities of pre-trained vision-language models to decode clinical information; 2) \textbf{Advanced text fusion techniques} including a cross-attention module and an Affine transformation fusion unit, to refine the conditioning process in cases where clinical information does not directly correspond to visual cues in images; and 3) \textbf{General implementation for GAN and diffusion models}. This research opens new avenues for employing deep generative models, surpassing traditional applications in data augmentation.
\section{Method}
\begin{figure}
    \centering
    \includegraphics[width=1\linewidth]{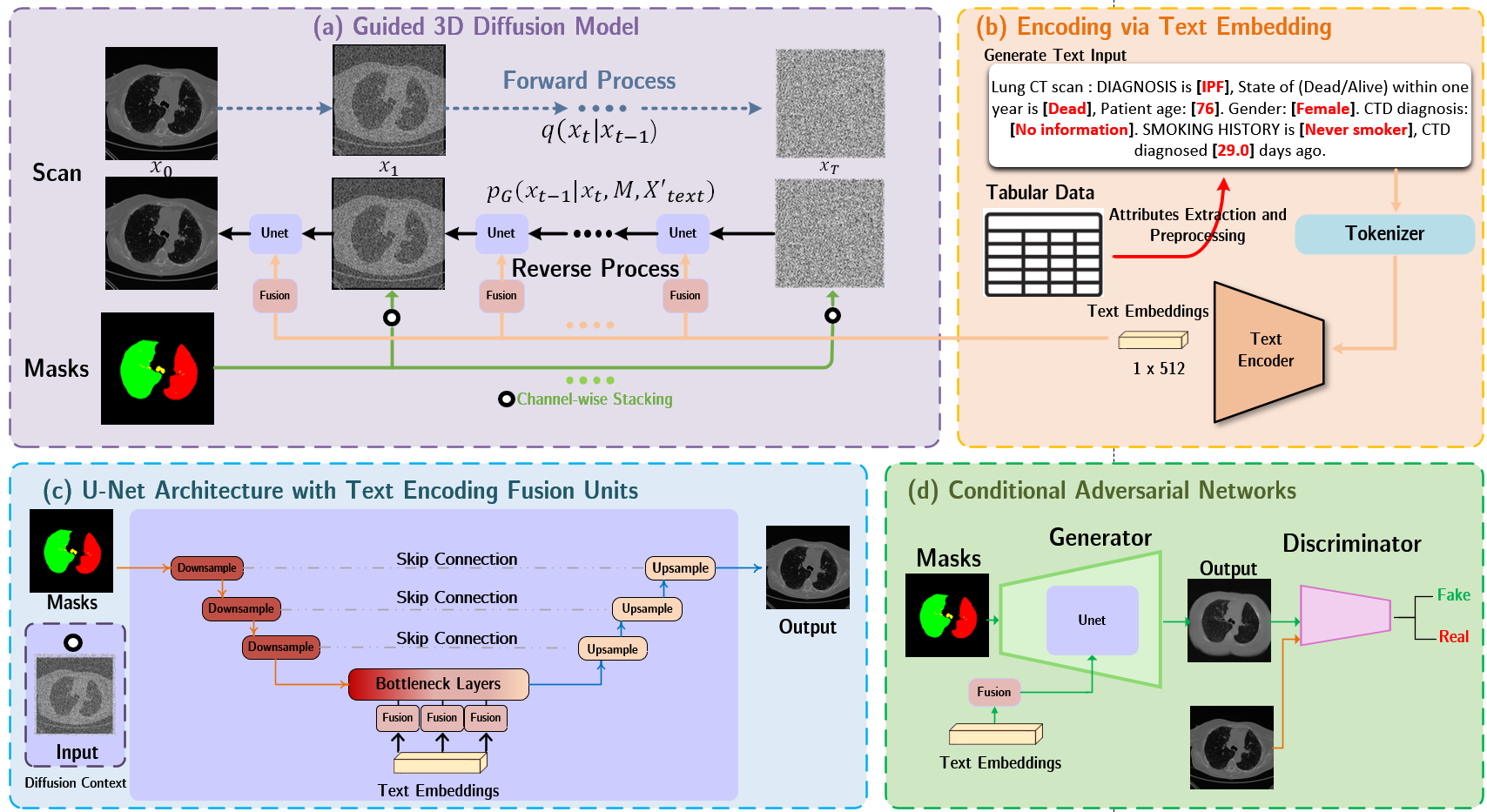}
    \caption{Overview of the proposed method. \textbf{Top left:} Illustration of the mechanism by which the text encoding embeddings are incorporated into the conditional diffusion model. \textbf{Top right:} Description of the process for utilizing tabular clinical data to obtain text embeddings from the pre-trained text encoder. \textbf{Bottom left:} A zoomed-in view of the fusion point where text embeddings integrate with the backbone of the models. \textbf{Bottom right:} Depiction of the compatibility of text fusion with a visual generator within the GAN framework.}
    \label{fig:method_pipeline}
\end{figure}

The general procedure, shown in \autoref{fig:method_pipeline} of our model pipeline proceeds as follows: First, we utilize any available tabular data related to lung CT scan masks, and use a transformation rule to normalize tabular data to text descriptions. We then employ a pre-trained Bert model \cite{alsentzer2019publicly}  specialized in the healthcare domain, to transform this tabular data into clinically relevant text descriptions. These text descriptions are fed into the frozen text encoder to obtain text embeddings. Next, the text embeddings are fused with the generative models using text-vision affine transformation fusion units.

 \begin{figure}
     \centering
     \includegraphics[width=1\linewidth]{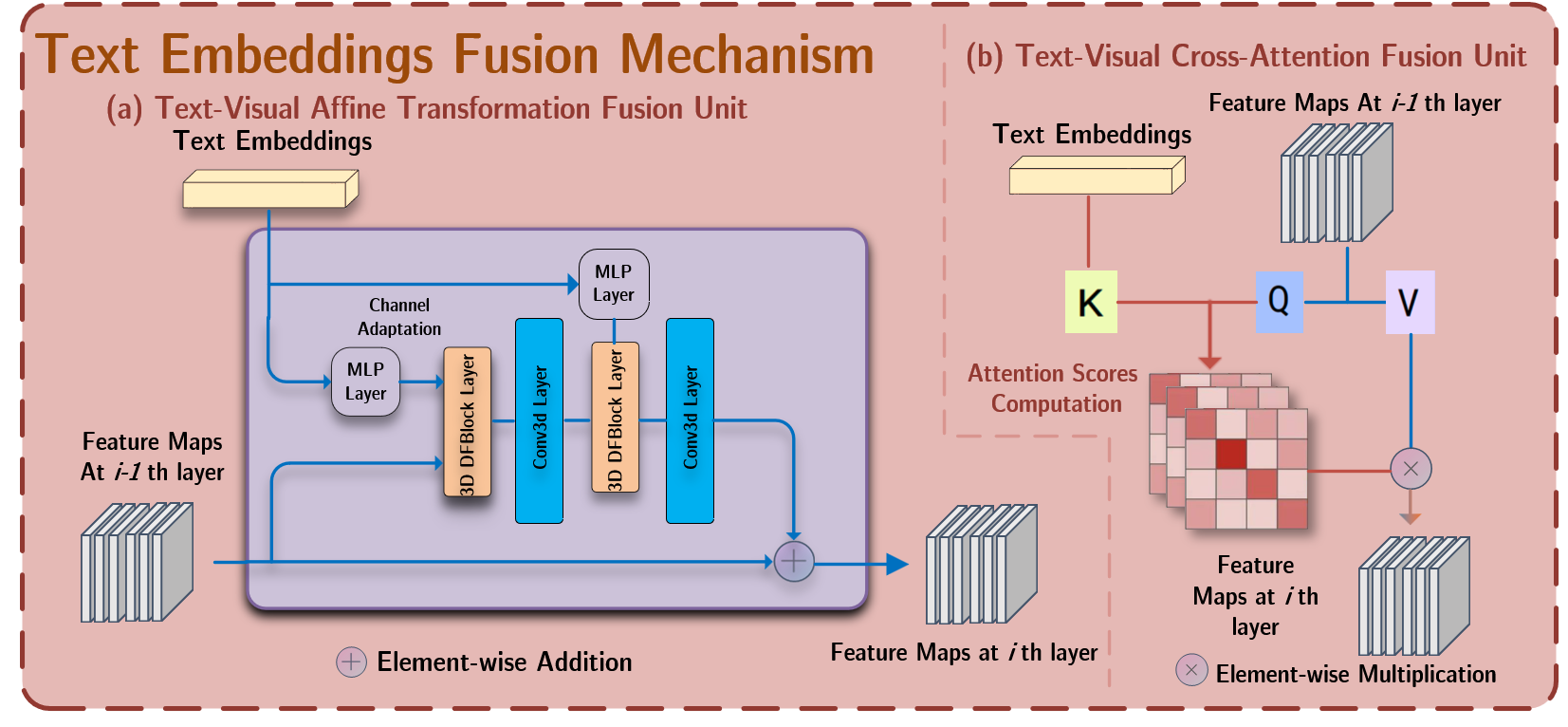}
     \caption{\textbf{Left}: Modified DFBlock Fusion Unit \textbf{Right}: Modified Cross-attention Fusion Unit}
     \label{fig:fusion_methods}
 \end{figure}

\subsection{Transformation of Tabular Data into Textual Representations} Electronic Health Record (EHR) data are predominantly stored in a tabular format. However, utilizing tabular data presents several challenges. The first issue is data missingness, leading to a reduction in the available data. The second issue is that tabular data cannot represent relationships between different classes. For instance, in diagnosing lung fibrosis, both Connective Tissue Disease-Interstitial Lung Disease (CTD-ILD) and Idiopathic Pulmonary Fibrosis (IPF) exhibit an Usual Interstitial Pneumonia (UIP) pattern. However, tabular data merely categorize these conditions into distinct classes (e.g., 0, 1) without acknowledging their similarities. Lastly, the CLIP model \cite{radford2021learning,wang2022medclip}  emerges as a notable method for generating text embeddings, offering a training-free approach for feature extraction. Considering these, we design a template that converts tabular data into textual descriptions. 

Our framework, outlined in \autoref{fig:method_pipeline} (b), processes each row of tabular data $x_i$ using a function $f$ to generate a textual description $x'_i$, following specific rules $R$. This involves filtering out attributes with unreasonable or missing values. Unlike other processes, we don't fill in missing values for text encoding with Bert, but \textbf{simply omitting them}. The process enriches text encoder context by concisely describing each data entry's attributes and their values. Mathematically, for the set of attributes $\mathcal{J}$, we form a matrix $X_{\text{text}}'$ for text embedding, with rows:
\begin{equation}
\forall i \in \{1, \ldots, |X|\}, x'_{i} = f([k, \mathbf{1}_{\{(x_{i})_{j} \neq \emptyset\}} ( x_{i}) ], R).
\end{equation}

\subsection{Incorporating Tabular Data as Text Embeddings in Generative Models}

Given that our method of dealing with the text embeddings from tabular data is adaptable across various generative models, we have showcased its effectiveness with two popular generative frameworks pix2pix \cite{isola2017image} and 3D diffusion models \cite{dhariwal2021diffusion,ho2020denoising}. We employ the schemes of text embedding fusions in two settings shown in \autoref{fig:method_pipeline} (a) and (d). While these fusion units are technically adaptable to any generative architecture, our choose different units for different generative backbones based on experimental observations.


\textbf{Text-Visual Affine Transformation Fusion Unit.} We enhance the training by incorporating an information mask with random noise during the denoising step and utilize $X_{\text {text }}^{'}$ for data synthesis. Adapting DFBlock from DF-GAN \cite{tao2022df} for 3D, we switch 2D convolutions to 3D and use an MLP linear layer for upsampling to maintain channel consistency for integrating text and visual features. This suits the U-shaped network's fusion needs, as shown in \autoref{fig:fusion_methods} (a).

For text embeddings, affine transformations use scaling ($\gamma$) and shifting ($\theta$) parameters to transform visual features, optimized via MLPs. Text embeddings are reshaped to align with visual feature channels before affine transformations:
\begin{equation}
    AFF(\boldsymbol{e_{i}} \mid x') = MLP_{\gamma}(x') \cdot e_{i} + MLP_{\theta}(x'),
\end{equation}
where MLPs match visual feature channels. This is followed by a 3D convolution and another fusion unit. Despite their efficiency in diffusion models, these units face modal collapse in Pix2pix, likely due to the original DFBlock's design for text-to-image synthesis, contrasting Pix2pix's conditional voxel generation.
\textbf{Text-Visual Cross-Attention Fusion Unit.} To address the problem identified earlier, we incorporate the conventional cross-attention mechanism within the Pix2pix method to enhance the integration of text embeddings. This approach is beneficial when there is no direct correlation between the textual information and structural guidance. We employ a tailored strategy that combines text embeddings with visual feature maps, where the text embeddings exclusively act as a "key" to selectively modulate the visual features as "query" and "value". This technique does not require a direct match between textual descriptions and visual conditions. Mathematically, shown in   \autoref{fig:fusion_methods} (b), consider an feature map denoted as $X \in \mathbb{R}^{B \times C \times D \times H \times W}$ and a text embedding vector $X_{text}^{'} \in \mathbb{R}^{B \times E}$, the tabular text embeddings cross-attention mechanism is defined as:

\begin{equation*}
\begin{aligned}
\text{Attention}(X, X^{'}_{text}) &= \text{softmax}\left(\frac{Q(X) \cdot K(X^{'}_{text}))^T}{\sqrt{d_k}}\right) \odot V(X), \\
Q(X) &= \text{Conv}_{Q}(X),\ K(X^{'}_{text})  = W_{K}X^{'}_{text},\ V(X) = \text{Conv}_{V}(X),\\
\end{aligned}
\end{equation*}
where $\odot$ denotes element-wise multiplication, $d_k$ is the scaling factor, typically the dimensionality of the key vectors,  $\text{softmax}$ is applied over the flattened spatial dimensions of the input tensor $X$ after being projected to the query space, $\text{Conv}_{Q}(\cdot)$ and $\text{Conv}_{V}(\cdot)$ are $1 \times 1 \times 1$ convolution operations to generate query and value representations, $W_{K}$ is a learnable weight matrix for the linear transformation of the text embedding into the key space.

\section{Experiments}

\subsection{Dataset and Implementation}
Our dataset is sourced from AIIB challenge, where the segmentation masks of lung regions and airways are available \cite{nan2023hunting}. We utilized training data of 50 samples, each with a size of $512 \times 512$ along the axial size, which is flexible depending on the individual images. To process the 3D image before training, we randomly crop the 3D image to a size of $256 \times 256 \times 64$ and apply a minimum number of mask criteria to ensure each crop contains some valid values of certain classes in the masks. For validation purposes, we preserve another set of 45 samples of the same size and evaluate the GAN metrics Fréchet Inception Distance (FID)\cite{NIPS20178a1d6947}, Kernel Inception Distance (KID)\cite{binkowski2018demystifying} , and Inception Score (IS)\cite{salimans2016improved} at a patch-wise level ($256 \times 256 \times 64$), with each test data randomly cropped 5 times.

For the details of the implementation, experiments were conducted on one A100 GPU for 1800 epochs, and the models were optimized through the Adam optimizer with an initial learning rate of $0.0001$ and $0.00001$ for the Pix2pix and DDpm methods, respectively, with both having a batch size of 2. The learning rate decayed after 800 epochs. The diffusion model is trained and evaluated with the timesteps of 250. The code implementation is based on \cite{wang2018high}.

\begin{figure}
     \centering
     \includegraphics[width=1\linewidth]{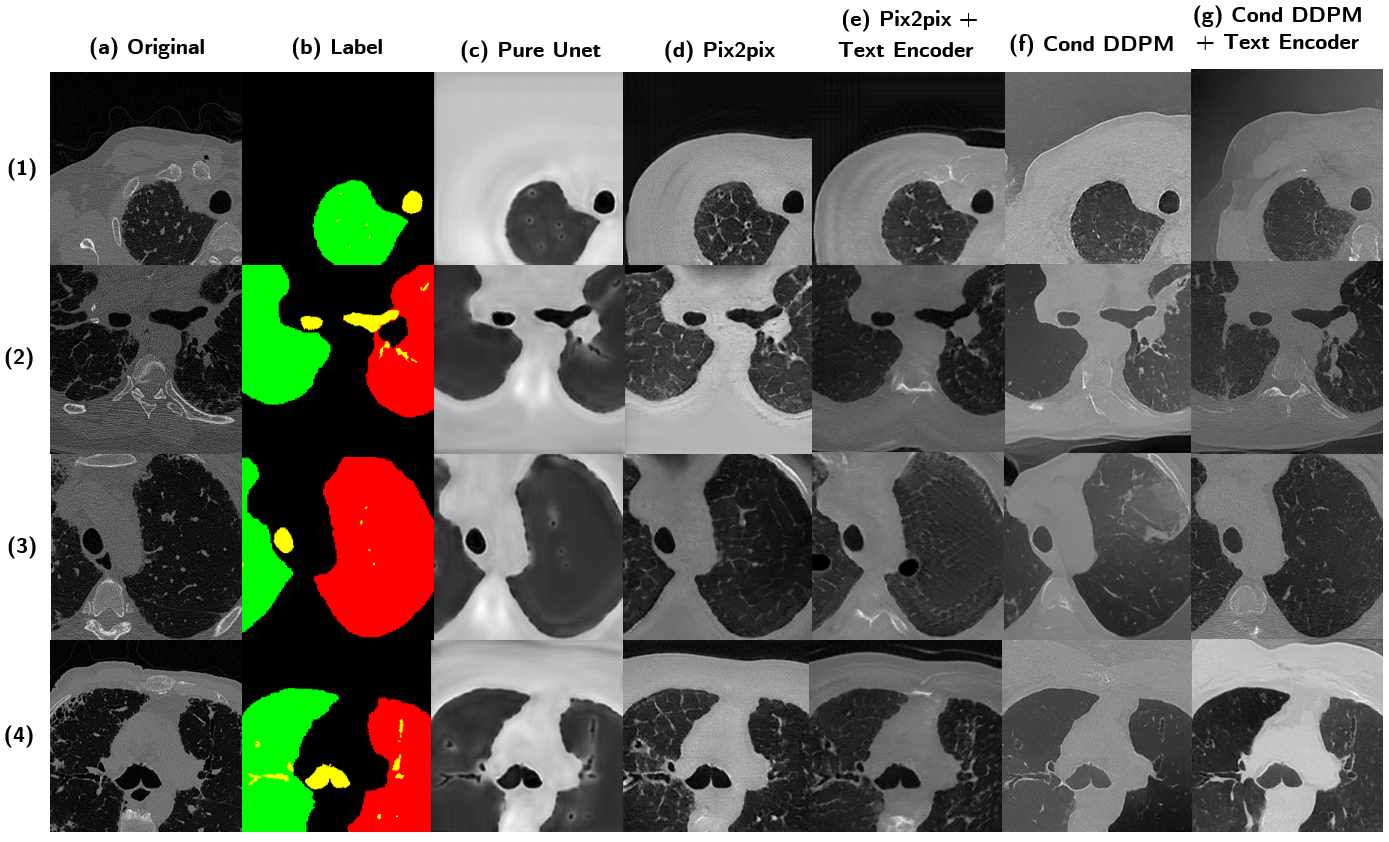}
    \caption{Visual comparison of reconstructed lung CT scans produced by four methods. From left to right: the original scan, followed by the four class labels—right lung, left lung, airway, and background. Subsequent columns present the results of Pure Unet, Pix2pix, Pix2pix  with Text Encoder, Cond DDPM and Cond DDPM with Text Encoder.}

     \label{fig:methodscomparsion}
 \end{figure}
 
\subsection{Synthesis Performance Comparison}
To evaluate the effectiveness of the tabular data utilization strategies with the existing deep learning framework, we choose the four models, 1) Pure UNet \cite{ronneberger2015u} 2) Pix2pix \cite{isola2017image},  3) Pix2pix + Text Encoder \cite{wang2022medclip} 4) Cond DDPM, 5) Cond DDPM + Text Encoder to perform the quantitative experiments evaluated on FID, KID and IS metrics. 

From the results in \autoref{table:resultscrossmethods} and \autoref{fig:methodscomparsion}, we observe the following trends. The Pix2pix method outperforms the other approaches in terms of FID and KID scores, underscoring the potential benefits of integrating tabular text embeddings within the existing conditional GAN framework. However, the performance gain is not replicated with the Conditional 3D Diffusion models when text embedding is introduced; a marginal decrease in performance is noted, with FID scores showing a 2\% reduction and KID scores dropping by 0.01. The nature of diffusion models, which rely on multiple iterative steps to reconstruct or generate images, may render text embeddings less effective as the incremental denoising process may diminish their impact. Nevertheless, the incorporation of tabular text embeddings still adds value.

Regarding the Inception Score (IS), our interest lies in the model's capacity to leverage the information provided to constrain the generation results. Observations indicate that methods incorporating text embeddings exhibit a decline in IS scores. This suggests that while the diversity of the synthesized output is constrained by the additional information, it is indicative that text embeddings are indeed influencing the behaviour of the models. This finding highlights the complexity of balancing the fidelity and diversity of generated images in generative models.

\begin{table}[]
    \caption{Comparative Analysis of Generative Models}
    \centering
    \resizebox{\linewidth}{!}{
    \begin{tabular}{|c|c|c|c|ccc|}
        \hline
        \multirow{2}{*}{\textbf{Models}} &  \multicolumn{3}{c|}{\textbf{Components}}& \multicolumn{3}{c|}{\textbf{Evaluation Metrics}}   \\
        \cline{2-7}
        & \textbf{Diffusion } & \textbf{GAN} & \textbf{Tabular Data}  & \textbf{FID} $\downarrow$ &\textbf{ KID} $\downarrow$ & \textbf{IS} $\uparrow$  \\
        \hline
        Pure Unet & \xmark&\xmark &\xmark & 221.99 
 & 0.207 (0.003)
 & 4.508(0.086)  \\  
        Pix2pix & \xmark &\cmark  &\xmark & 
  143.779
 & 0.098 (0.002)
 & 3.920(0.054) \\
        Pix2pix + Text Encoder & \xmark & \cmark &\cmark   & 113.097

 &  0.077 (0.002)
 &  3.552(0.054)\\ 
        Cond DDPM &\cmark  & \xmark & \xmark & 160.0576 
 &  0.114 (0.002)
 &  3.721(0.044)  
\\ 
        Cond DDPM + Text Encoder & \cmark &\xmark  &\cmark   & 163.0374 
 & 0.120 (0.003)
 &  3.136(0.128)
 \\
        \hline
\end{tabular}}

    \label{table:resultscrossmethods}
\end{table}

\subsection{Pattern Identification Analysis}
\begin{figure}
     \centering
     \includegraphics[width=1\linewidth]{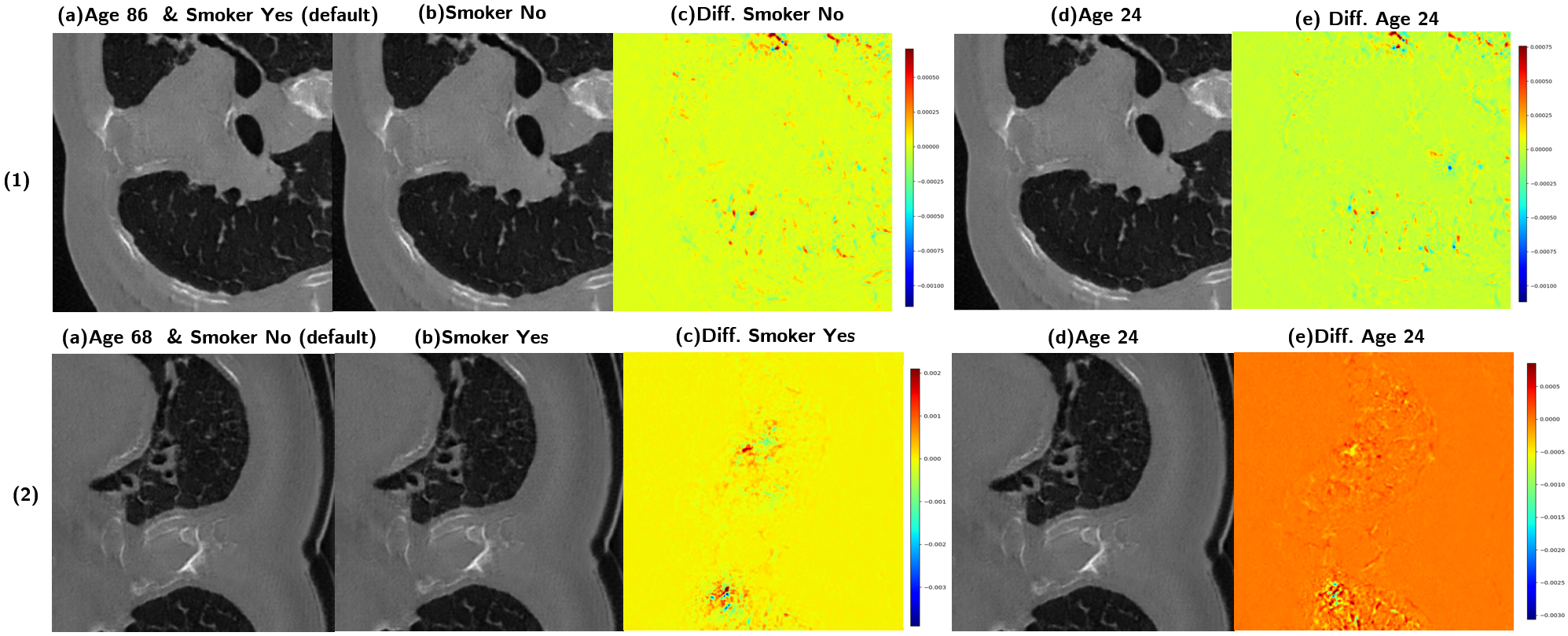}
     \caption{Qualitative Assessment of Pixel-Level Variations Following Prompt Modification: Illustrative Cases Demonstrating the Impact of Altered Prompt Content on Prediction Outcomes.}
     \label{fig:absolute_difference}
\end{figure}

We provide visual comparisons to show how clinical data influences image generation. A control experiment examined the impact of slight changes in tabular text descriptions on CT scan synthesis, measuring change by the difference in voxel values, using the Pix2pix and Text Encoder method. As shown in \autoref{fig:absolute_difference}, heatmaps highlight differences in test samples with varying "age" and "smoker" descriptions.

Transitioning from "non-smoker" to "smoker" status resulted in a distinct intensity pattern, with dot-like increases linked to the formation of lung nodules or inflammation and dot-like decreases associated with the destruction of lung tissue and the formation of air spaces. This is observed in \autoref{fig:absolute_difference} (c2), aligning with clinical evidence that smoking can lead to a dual effect on lung density: increased in areas of tissue densification and decreased where lung tissue is compromised. Overall, an increase in the intensity is consistently observed. 

Conversely, we did not identify a consistent trend in intensity changes with age variations, which we attribute to the limitations of our dataset, starting at a minimum age greater than 30, and the possible inability of language models to discern numerical relationships. Future efforts will focus on independently integrating numerical inputs to overcome this challenge.

This observation confirms that text encoders such as BERT or CLIP are adequately sensitive to condition the synthesis through their integration within the generative framework, employing mechanisms like cross-attention or affine transformation fusion.

\section{Conclusion} 

In this study, we developed a versatile framework demonstrating the potential of deep generative models for uncovering invisible patterns in medical images associated with various clinical states. We innovatively transformed tabular data into textual descriptions, enabling the integration of clinical data with image synthesis methods through pre-trained vision-language models. Given the unique optimization challenges of generative models, we designed two distinct units to fuse textual and structural guidance for both GAN and Diffusion Model backbones, ensuring high-quality image synthesis while maintaining clinical relevance.

Moreover, we observed that while language models are good at understanding abstract concepts like life and death, they struggle with numerical understanding, such as recognizing that age 24 is less than age 68, which suggests a need for inputs beyond just text. In our future work, we aim to explore various conditioning to broaden the use of generative models beyond only data augmentation.

Our results are promising, showing that generative models can identify unseen patterns related to specific clinical attributes, such as the smoking status in lungs. These findings underscore the potential for significant advancements in the early detection of lung diseases and other complex medical conditions. This research opens new avenues for employing deep generative models, surpassing traditional applications in data augmentation.

\bibliographystyle{plain}
\bibliography{mybibliography}

\end{document}